\def\year{2019}\relax
\documentclass[letterpaper]{article} 
\usepackage{aaai19}  
\usepackage{times}  
\usepackage{helvet} 
\usepackage{courier}  
\usepackage[hyphens]{url}  
\usepackage{graphicx} 
\usepackage{graphicx}  
\usepackage{multirow}

\title{Attend to the beginning: A study on using bidirectional attention for extractive summarization}
\author{Ahmed Magooda\textsuperscript{\rm 1}\thanks{This work was done during internship}, \Large \textbf{ Cezary Marcjan}\textsuperscript{\rm 2}\\
\textsuperscript{\rm 1} University of Pittsburgh, Pitssburgh, PA, USA\\
\textsuperscript{\rm 2} Microsoft Research FUSE lab, CCP, Bellevue, WA, USA\\
aem132@pitt.edu, cezarym@microsoft.com
}

\begin{document}

\maketitle

\begin{abstract}
Forum discussion data differ in both structure and properties from generic form of textual data such as news. Henceforth, summarization techniques should, in turn, make use of such differences, and craft models that can benefit from the structural nature of discussion data. In this work, we propose attending to the beginning of a document, to improve the performance of extractive summarization models when applied to forum discussion data. Evaluations demonstrated that with the help of bidirectional attention mechanism, attending to the beginning of a document (initial comment/post) in a discussion thread, can introduce a consistent boost in ROUGE scores, as well as introducing a new State Of The Art (SOTA) ROUGE scores on the forum discussions dataset. Additionally, we explored whether this hypothesis is extendable to other generic forms of textual data. We make use of the tendency of introducing important information early in the text, by attending to the first few sentences in generic textual data. Evaluations demonstrated that attending to introductory sentences using bidirectional attention, improves the performance of extractive summarization models when even applied to more generic form of textual data.

\end{abstract}

\section{Introduction}
Recently, automatic text summarization models either extractive or abstractive witnessed fast performance strides due to the emergence of deep learning models specially seq2seq models. A large number of recent neural abstractive summarization models employ the encoder-decoder structure \cite{see2017get,gehrmann2018bottom,paulus2017deep} to convert the input sequence into a relatively shorter sequence. Most of the recent extractive models, on the other hand, employ only an encoder part to convert the input sequence into a fixed feature vector, followed by a classification part \cite{nallapati2017summarunner,liu2019text}. Text summarization has been applied to different natural language domains; news, academic papers, emails, meeting notes, forum discussions, etc.. While some models can be transferable from one domain to the other, it might be more beneficial to craft additional modifications in those models to account for differences between domains. Forum discussion data \cite{tarnpradab2017toward}, for example, is different in both structure and properties when compared to generic textual data such as news. Discussion threads usually start with an initial post/comment (i.e. seeking knowledge, or help, etc..). The following comments tend to target the initial post/comment, providing additional information or opinions. With that said, a question arises. Can we enhance existing summarization models to benefit from such properties ?

Inspired by \cite{seo2016bidirectional} we propose integrating bidirectional attention mechanism in extractive summarization models, to help to attend to early pieces of text (initial comment). The main objective is to benefit from the dependency between the initial comment and the following comments and try to distinguish between important, and irrelevant or superficial replies. Moreover, recent research by \cite{jung2019earlier} showed that in some domains, humans tend to introduce relatively important information early at the beginning of textual articles. Unlike discussion threads, We explore the benefit of attending to the beginning in a more generic textual setting. Simply by integrating bidirectional attention mechanism and attending to the first few sentences in a document. We conducted some experiments to evaluate this hypothesis using a dataset of generic (non-discussion based) documents. Thus our contributions in this work are three-fold. First, we introduce integrating bidirectional attention mechanism into extractive summarization models, to help to attend to earlier pieces of text. Second, we achieved a new SOTA on the forum discussion dataset through the proposed attending to the beginning mechanism. Third, to further verify the transferability of our hypothesis (i.e. attending to the beginning), we perform evaluations to show that attending to earlier sentence in a more generic text, can also benefit summarization models. on different domains other than discussions.

\section{Related Work}
Automatic text summarization has seen increasing interest and improved performance due to the emergence of seq2seq models \cite{sutskever2014sequence} and attention mechanisms \cite{bahdanau2014neural}. This is true for both automatically generating coherent summary (abstractive summarization), and extracting salient pieces of text (extractive summarization). The majority of recent research has been directed towards the news domain \cite{see2017get,paulus2017deep} (i.e. due to the existence of huge annotated datasets CNN/DailMail, Gigawords, Newyork Times). Unlike news, other domains such as (emails, discussions, meeting notes, students feedback, and opinions) can still be considered underexplored. 

Recent efforts to tackle such domains started to emerge, \cite{luo2016improved} targeted student feedback summarization by extracting a set of representative phrases. \cite{li-etal-2019-keep} proposed doing abstractive summarization for meeting notes by employing textual and visual information into a multi-model setting. \cite{li2019towards,yang2018aspect} tackled the problem of opinion and review summarization. A work targeting similar domain as ours is done by \cite{tarnpradab2017toward}. In which they proposed doing hierarchical attention to perform extractive summarization over a dataset of forum discussions collected from trip advisor. Another work which shares a similar design concept as ours was done by \cite{wang2019biset}. They also integrated bidirectional attention mechanism in their model, however, their model and ours are different in both intuition and application. The major motive to integrate bidirectional attention in their design is to attend to an external template during the summarization process, while in ours we propose attending to early pieces of input text using the bidirectional attention mechanism. Another major difference is the intended application. They developed their model for the task of abstractive summarization over news data, while ours is intended for extractive summarization task.

\section{Dataset}
In this work, we employ two extractive summarization datasets. First, we used the discussion dataset proposed by \cite{tarnpradab2017toward}\footnote{Data can be downloaded from https://www.dropbox.com/s/heevii01b1l6s0a/threadDataSet.zip?dl=0}. The discussion dataset is extracted from trip advisor forum discussions. The data consists of 700 threads. In their work,  \cite{tarnpradab2017toward} used 600 threads for training and 100 for validation. We didn't use the same data distribution reported by the authors, however, we kept the same testing data size for comparability reasons. We used our own split to verify the utility of our proposed techniques. We used 500 threads for training, 100 for validation, and 100 for testing. Moreover, we conducted additional experiments using (MSW) dataset\cite{jha2020artemis}\footnote{This dataset is not publicly available}. MSW dataset is a generic textual dataset that is used to verify the transferability of our hypothesis to more generic textual domains. We verify whether we can benefit from documents' structure, and human's tendency to present important information earlier, by attending to early sentences. MSW dataset consists of a collection of 532 generic documents of different domains. We split the data into training, validation, and testing of 266, 138, and 128 documents respectively. 
\begin{table}[t]
\begin{center}
\small
\begin{tabular}{|c|c|c|c|}
\hline \textbf{Dataset} & \textbf{Part} & \textbf{\# Documents} & \textbf{Total \# Sentences}\\
\hline
\multirow{3}{*}{\bf Trip Advisor} & Train & 500 & 29671\\
\cline{2-4}
& Val & 100 & 6251\\
\cline{2-4}
& Test & 100 & 4280\\
\cline{2-4}
\hline  \multirow{3}{*}{\bf MSW} & Train &266 & 19748\\
\cline{2-4}
& Val &138 & 11488\\
\cline{2-4}
& Test &128 & 9898\\
\cline{2-4}
\hline
\end{tabular}
\end{center}
\caption{\label{tab:datasets} Model Datasets}
\end{table}
Table \ref{tab:datasets} summarizes the distribution of datasets used.

\section{Baselines}
In order to validate our hypothesis and show the utility of our proposed enhancements, we implemented 4 baselines.
The following sections provide additional details regarding each of the baselines implemented.

\subsection{Sumy\footnote{https://pypi.org/project/sumy/}}
We used the python off the shelf package for text summarization Sumy. Summarization is done hierarchically, where each comment from the discussion thread is passed separately to Sumy. The resultant summaries are then combined and passed to Sumy as one final document to get the thread summary.

\subsection{LSA + clustering}
We implemented a simple baseline for extractive summarization. The baseline uses Latent Semantic Analysis (LSA) to embed sentences into vector space. Sentences are then clustered using the K-means clustering algorithm. We use a number of clusters = $\sqrt{n}$ where $n$ is the number of sentences in the input document. Lastly, for each cluster, a cluster head is picked. The cluster head is the sentence closest to the mean point of the cluster. 

\subsection{SummaRuNNer}
SummaRuNNer is an auto-regressive extractive summarization model proposed by \cite{nallapati2017summarunner}\footnote{Refer to paper for original model design}. 

\subsection{SiATL} 
(SiATL) is sentence classification model developed by \cite{chronopoulou2019embarrassingly}. The model employs multi-task learning by integrating language modeling auxiliary loss during the training process. SiATL model was developed originally as a sentence classification model. However, we decided to deal with extractive summarization as a pure sentence classification problem, and use SiATL model as extractive summarizer.

Unlike SummaRuNNer which is an auto-regressive model (i.e. previous decisions made by the model, affect its future decisions), SiATL performs classification independently for each sentence. While it may always seem that autoregressive models would perform better, \cite{xiao2019extractive} showed that non-auto-regressive models can sometimes be more efficient. Thus, we decided to use the SiATL model as a baseline, and compare the performance of auto-regressive and non-auto-regressive models within the extent of our study.

\section{Attend to the beginning}
Throughout this work, we hypothesize that attending to the initial part of a text during extractive summarization would help in selecting more salient sentences. The intuition is that in some situations(e.g. discussion threads), the initial part of a text holds important topical information. Henceforth it renders an important factor in selecting salient sentences for summarization objective. Thus we validate this hypothesis by calculating the importance of a sentence with respect to the initial part of the text, in the form of attention. Influenced by \cite{seo2016bidirectional,wang2019biset}, the same interaction approach is employed here to produce beginning-aware sentence representations, for each sentence in the document. First, a sentence representation is produced for each sentence of the document as well as each sentence of the beginning part of the document (i.e. initial post/comment in the case of discussion dataset). Similarity matrix $S \in R^{m\times n}$ is then computed for each pair of document and beginning part sentences, $s_{ij}$ = $W_0[h_{i}^{d};h_{j}^{b};h_{i}^{d}\otimes h_{j}^{ip}]$. Where $n$ is the number of sentences in the beginning part, $m$ is the number of sentences in the input document, ';' is the concatenation operator, $h_{i}^{d}$ is the sentence representation of the i's  sentence in the document, and $h_{j}^{b}$ is sentence representation of the j's sentence in the beginning part. Each row and column of $S$ is then normalized by softmax, which produces two new matrices $\overline{S}$ and $\overline{\overline{S}}$. Bidirectional attention is then calculated as $A$ = $\overline{S}\cdot h^{b}$, $B$ = $\overline{S}\cdot \overline{\overline{S}}^T\cdot h^{b}$, where $A$ represents  document-to-beginning attention and $B$ represents beginning-to-document attention. Finally, we obtain the beginning-aware sentence representations for each sentence in the document: $G_{i}^{d}$  $\forall$ $i\in m$, where \quad $G_{i}^{d}$ = $[h_{i}^{d};A_i;h_{i}^{d}\otimes A_i;h_{i}^{d}\otimes B_i]$ 

The underlying mechanism to integrate bidirectional attention in (SiATL, and SummaRuNNer)\footnote{Code is available at github.com/amagooda/SummaRuNNer\_coattention} is very much the same, except for the level of granularity in which attention operates on. SummaRuNNer operates on the level of document, so the bidirectional attention mechanism is calculated on the level of sentences between (all document sentences, and the beginning sentences) (i.e. $h_{i}^{d}$ is the sentence representation of the i's  sentence in the document, and $h_{j}^{b}$ is sentence representation of the j's sentence in the beginning part). On the other hand, SiATL operates on the level of sentence, thus bidirectional attention is calculated between words of the input sentence, and words of the beginning part of the document (i.e. $h_{i}^{d}$ is the word representation of the i's  word in the input sentence, and $h_{j}^{b}$ is word representation of the j's word in the beginning part).





\section{Additional Proposed Modifications}
\subsection{BERT Embedding}
Recently released BERT\cite{devlin2018bert} embeddings showed the ability to outperform simply using shallow word embeddings. Moreover, It helped pushing the state of the art for numerous tasks within the NLP community. In this work, we integrate BERT embeddings within the SummaRuNNer model. Instead of initializing word embeddings randomly, BERT embeddings are used. The model embedding layer is initialized with Bert embedding and froze during the training phase.

\subsection{Keyword Extraction}
Attention mechanisms aim to weight tokens differently based on their importance.  Another modification we introduce in this work is directed towards feeding the model with an extra signal. The extra signal, in this case, is keywords. The intuition behind feeding the model with keywords is pushing the model to give more attention to some specific words. The way we integrate keywords in SummaRuNNer model is by extracting keywords from each sentence $S_i$. Then separately encode the keywords  into hidden states using BiLSTM ($h_j^{kwi}$ $\forall \quad j \in N_{kwi}$, where $N_{kwi}$ is the number of extracted keywords from sentence $S_i$). The last hidden state is then used to represent all the keywords ($h_i^{kw}$). A new sentence embedding ($h_i^{dkw}$) is then formed by directly concatenating the original document aware sentence representation and the keywords representation.
$$h_i^{dkw} = [h_i^{d} ; h_i^{kw}]$$

\section{Experiments}\label{sec:experiments}
To verify our hypotheses and validate the utility of our proposed modifications, we conducted a number of experiments. Our experimental designs address the following hypotheses:\\
\textbf{Hypothesis 1 (H1)} : Attending to the beginning of a discussion thread, would help  extractive summarization models to select more salient sentences.\\
\textbf{Hypothesis 2 (H2)} : Non-auto-regressive models such as SiATL might be more suitable for thread discussion summarization, compared to auto-regressive models such as SummaRuNNer.\\
\textbf{Hypothesis 3 (H3)} : Adding additional features, such as contextual embeddings (e.g. BERT) and keywords can give summarization models a boost in performance.\\
\textbf{Hypothesis 4 (H4)} : Attend to the beginning is transferable to different forms of text other than discussion threads.\\
\\
\textbf{LSA + Kmeans}. As part of the LSA baseline, two LSA vector spaces were used; First a vector space trained on a part of Wikipedia. Second, a vector space trained using the forum discussion dataset. ScikitLearn python package was used to produce LSA vector spaces of 200 dimensions. 
\\\\
\textbf{SummaRuNNer}. We implemented SummaRuNNer model following \cite{nallapati2017summarunner}. To operate on forum discussion data, comments are split into sentences using Stanford sentence parser. All sentences are then concatenated into a single document.\\ $D$ = \{$S_i$ for $S_i$ in $C_{1}$\} ; .. ; \{$S_i$ for $S_i$ in $C_{j}$\}  for $j\in [1..n]$\\ Where 
$S_i$ is the i's sentence, ';' is the concatenation operator, $C_j$ is the j's comment, and $n$ is the number of comments. Moreover, to operate on MSW dataset, each document is also split into sentences using Stanford sentence parser. SummaRuNNer used randomly initialized embeddings of size 64. The hidden state size of the LSTM is 128. Input sentences are truncated to 75 tokens, while shorter sentences are padded.  The model is trained with batch size = 32 for 100 epoch. We calculate ROUGE score over the development set on each epoch. Later on, the checkpoint with maximum ROUGE is used for testing.
\\\\
\textbf{SiATL (H2)}. We used the implementation of SiATL released by the authors\footnote{https://github.com/alexandra-chron/siatl}. The model used embeddings of size 400 dimensions. The hidden state size of the shared LSTM is 1000. The task LSTM is of size 100. Input sentences are truncated to 80 tokens, while shorter sentences are padded. The model is trained with batch size = 32 for 100 epoch. Similarly, we calculate the ROUGE score over the development set. Later on, the checkpoint with the maximum ROUGE score is used for testing.
\\\\
\textbf{SummaRuNNer + Bidirectional Att. (H1, and H4)}. The bidirectional attention mechanism integrated in SummaRuNNer operates on the level of document. To conduct experiments on forum discussion data, the beginning part is the first comment which is the initial post in the thread. On the other hand, during experiments on the MSW dataset, the beginning part is the first $N$ sentences in each document. In this work, we used $N=3$.
\\\\
\textbf{SummaRuNNer + BERT Embedding (H3)}. To initialize SummaRuNNer with BERT word embeddings, BERT base uncased embeddings were used\footnote{https://storage.googleapis.com/bert\_models/2018\_10\_18/uncased\_L-12\_H-768\_A-12.zip}. Each word is represented by the concatenation of BERT's last two layers, which leads to a word representation of size = 2x768 = 1536. We tried combining different number of layers(1, 2, and 3) for each word representation. We found that combining 2, or 3 layers performs better than using only the last layer. We decided to use only 2 layers to reduce the number of model parameters.
\\\\
\textbf{SummaRuNNer + Keyword extraction (H3)}. To extract keywords, we use Rapid Automatic Keyword Extraction  (RAKE) \cite{rose2010automatic} to identify keywords. For each sentence in the document, Keywords extracted and concatenated. Each pair of sentence and corresponding concatenated keywords are then passed to SummaRuNNer as separate inputs. 
\\\\
\textbf{SiATL + Bidirectional Att. (H1, H2, and H4)}. Unlike SummaRuNNer, SiATL operates on the level of individual sentences. Thus, the bidirectional attention mechanism integrated operates on the level of words. To conduct experiments on forum discussion data, the beginning part is all the words from the initial comment in the thread. On the other hand, during experiments on the MSW dataset, the beginning part is all the words from the first $N$ sentences in each document. In this work, we used $N=3$.

\section{Results on forum dataset} \label{sec:forum_scores}
Table \ref{tab:forum_results} presents  summarization performance results for the 2 non-neural extractive baselines,  for the original and proposed variants of the two summarization models SummaRuNNer and SiATL, and finally for the highest score reported by \cite{tarnpradab2017toward}. Following \cite{tarnpradab2017toward} and other recent work,  performance is evaluated using  ROUGE-1 (R-1), ROUGE-2 (R-2), and ROUGE-L (R-$L$) \cite{lin2004rouge} on F1.

\begin{table}[!ht]
\begin{center}
\small
\begin{tabular}{|l|c|c|c||c|}
\hline
\textbf{Summarization Model} & \textbf{R-1}  & \textbf{R-2}  & \textbf{R-L} & \\
\hline
\multicolumn{5}{|c|}{\textbf{Baselines}}\\
\hline
Tarnpradab (Best) & 37.6 & 14.4 & 33.8 & 1\\
\hline
Sumy  & 38  & 15.06  & 21.95 & 2\\
\hline
LSA + kmeans (Discussions) & 35.94 & 19.05  & 23.03 & 3\\
\hline
LSA + kmeans (Wikipedia) & 35.4 & 18.49  & 22.57 & 4\\
\hline
SummaRuNNer (Basic) & 36.97 & 15.84 & 24.5 & 5\\
\hline
\underline{SiATL (Self Attention)} & \underline{45.15}  & \underline{26.12} & \underline{43.3} & 6\\
\hline
\hline

\multicolumn{5}{|c|}{\textbf{SummaRuNNer}}\\
\hline
 + Bidir. Att.  & 37.46 &  16.17 & 24.5 & 7\\
\hline
 + BERT & 38.48 & 16.88 & 25.63 & 8\\
\hline
 + Keywords (KWs) & 37.3 & 15.85 & 24.98 & 9\\
\hline
 + Bidir. Att.  + KWs & 37.79 & 16.25 & 24.76 & 10\\
\hline
 + BERT + KWs & 37.97 & 16.75 & 25.85 & 11\\
\hline
 \underline{+ BERT +  Bidir. Att. } & \underline{39.36} & \underline{17.71} & \underline{26.78} & 12\\
\hline
 + BERT +  Bidir. Att.  + KWs & 38.43 & 16.74 & 25.65 & 13\\
\hline
\hline

\multicolumn{5}{|c|}{\textbf{SiATL}}\\
\hline
\textbf{\textit{\underline{Bidir. Att.}}} & \textbf{\textit{\underline{46.5}}} & \textbf{\textit{\underline{28.53}}} & \textbf{\textit{\underline{44.65}}} & 14\\
\hline
 Self Att.+ Bidir. Att.  & 46.32 & 28.69 & 44.41 & 15\\
\hline
\end{tabular}
\end{center}
\caption{\label{tab:forum_results} ROUGE results. \textit{Italics}  indicates outperforms all baselines. \textbf{Boldface} indicates  best result over all models. \underline{Underlining} indicates  best result within model group}
\end{table}
The motivation for using bidirectional attention mechanism is our hypothesis \textbf{(H1)}. Table ~\ref{tab:forum_results} supports this hypothesis. All ROUGE scores for SummaRuNNer and SiATL, that involves attending to the beginning by using bidirectional attention mechanism (rows 7, 10, 12, and 14), Outperform their corresponding counterpart, without using bidirectional attention (rows 5, 9, 8, and 6) respectively. Our second hypothesis \textbf{(H2)} is non-auto-regressive models might be more suitable than auto-regressive ones, for discussion summarization. Table \ref{tab:forum_results} shows that using non-auto-regressive model (SiATL) indeed improve ROUGE scores compared to the auto-regressive model (SummaRuNNer). In rows 6 and 5, we see that SiATL improved R-1 scores from 36.97 to 45.15. Similarly, R-2 and R-$L$ are also improved from 15.84 to 26.12 and from 24.5 to 43.3 respectively. Additionally, SiATL introduced a new SOTA, with a huge improvement in ROUGE scores compared to the previous work using hierarchical attention (rows 6, 14 and 1), in which R-1 improved by 23.6\%. R-2 improved by 98.12\%, and finally, R-$L$ improved by 32.1\%. We also see the same benefits of attending to the beginning for SiATL model: compared to using only the self-attention mechanism (i.e. original model), using only bidirectional attention or combining both attention mechanisms (self and bidirectional) boost ROUGE scores (rows 6, 14, and 15)

Our next hypothesis \textbf{(H3)} is that enriching models with additional features such as (Contextual embeddings, or keywords) would boost the performance. For these experiments, we only used SummaRuNNer model, since it still has a room for improvement to catch up with the SiATL model. Table~\ref{tab:forum_results} shows that our third hypothesis is a valid one, but not for all cases. It shows that while adding Contextual embeddings by itself, or adding keywords by itself helps the model. Combining contextual embeddings with keywords tends to harm the model. We can see that Adding keywords to both variants of SummaRuNNer (original, and with bidirectional attention) introduces a slight improvement over ROUGE scores (rows 5, 7 and 9, 10). Where R-1 improved from (36.97, and 37.46) to (37.3, and 37.79) respectively. R-2 improved from (15.84, and 16.17) to (15.85, and 16.25), and R-$L$ improved from (24.5, and 24.5) to (24.98, and 24.76). Similarly, adding BERT contextual embedding introduces a good improvement over ROUGE scores for both variants of SummaRuNNer (rows 5, 7 and 8, 12). Where R-1 improved from (36.97, and 37.46) to (38.48, and 39.36) respectively. R-2 improved from (15.84, and 16.17) to (16.88, and 17.71), and R-$L$ improved from (24.5, and 24.5) to (26.99, and 27.95). Surprisingly, adding both features (BERT, and keywords), tends to be harmful to the model (rows 8, 12 and 11, 13). Further analysis is still needed to reach a solid conclusion for such behavior. 

\section{Results on MSW dataset}
Table \ref{tab:msword_results} presents summarization performance results for 
the original and proposed variants of SummaRuNNer, for the best-performing variant of SiATL.
The motivation behind conducting experiments on the MSW dataset is to validate \textbf{our last hypothesis (H4)}. We can see that table \ref{tab:msword_results} clearly shows that our hypothesis is a valid one. It shows that attending to the beginning of a document helps selecting more salient sentences, not just for discussion threads, but even for generic textual documents. Similar to the results on the discussions dataset, we can see that attending to the beginning through a bidirectional attention mechanism boosts ROUGE scores (rows 1, and 2). Additionally, we can see that combining bidirectional attention with BERT embeddings further improves the performance (rows 1, and 4).

\begin{table}[ht!]
\begin{center}
\small
\begin{tabular}{|l|c|c|c||c|}
\hline
\textbf{Model} & \textbf{R-1}  & \textbf{R-2}  & \textbf{R-L} & \\
\hline
SummaRuNNer & 63.48  & 54.51  & 61.66 & 1\\
\hline
\textit{+ Bidir. Att.} & \textit{64.23} & \textit{55.23} & \textit{62.21} & 2\\
\hline
\textit{+ BERT} & \textit{65.81} & \textit{57.99} & \textit{63.9} & 3\\
\hline
\textbf{\textit{+ Bidir. Att. + BERT}} & \textbf{\textit{66.12}} & \textbf{\textit{58.56}} & \textbf{\textit{64.48}} & 4\\
\hline
SiATL (Self Att. + Bidir. Att.) & 44.81 & 27.02 & 42.79 & 5\\
\hline
\end{tabular}
\end{center}
\caption{\label{tab:msword_results} ROUGE results over MSW dataset.}
\end{table}

\section{Discussion \& Analysis}
Unlike its promising performance on discussions dataset (table \ref{tab:forum_results}), SiATL performed poorly on MSW dataset (table \ref{tab:msword_results}). Surprisingly, it was only able to outperform the lead3 baseline. Through analyzing different criteria of the generated output for SummaRuNNer and SiATL, over the two used datasets. We observed that SiATL tends to generate longer summaries compared to SummaRuNNer, and this most likely due to its non-auto-regressive nature. SummaRuNNer, on the other hand, tends to generate shorter summaries. Table \ref{tab:analysis} shows the average and standard deviation of the number of sentences generated using SummaRuNNer and SiATL model, compared to the human annotation. It shows that for the forum discussion dataset, the expected summary length is $\sim$ 14 sentences. For the same dataset, the SiATL model generates summaries of length $\sim$ 16 sentences, while SummaRuNNer generates summaries of length $\sim$ 8 sentences. This can justify the superior performance of SiATL compared to SummaRuNNer on the forum discussion dataset. On the other hand, we can notice that the expected summary length for the MSW dataset is $\sim$ 8 sentences. For the same dataset, SummaRuNNer consistently generates shorter summaries compared to SiATL of lengths $\sim$ 6.5 compared to 22 respectively. It is clear that the huge difference in the length of the summary between the human and SiATL generated is the reason SiATL underperforms on the MSW dataset. A potential solution for the SiATL model would be by adding a final post-processing step (e.g. clustering, redundancy reduction, etc..). The rule of the post-processing step would be slightly filtering the generated summary, and help to pick a number of sentences close to the average number humans select.

\begin{table}[!ht]
\begin{center}
\small
\begin{tabular}{|l|c|c||c|c|}
\hline
\textbf{Model} & \multicolumn{2}{|c||}{\bf Forum Discussions} & \multicolumn{2}{|c|}{\bf MSW}\\
\cline{2-5}
 & \textbf{Avg}  & \textbf{Std}  & \textbf{Avg} & \textbf{Std} \\
\hline
Human & 13.38 & 8.16 & 7.15 & 7.58\\
\hline
SummaRuNNer & 8.2 & 3.52 & 6.4 & 3.6\\
\hline
SiATL & 16 & 6.48 & 21.85 & 12.69\\
\hline
\end{tabular}
\end{center}
\caption{\label{tab:analysis} Average and standard deviation of the number of sentences generated from each model, and the human selected sentences}
\end{table}

\section{Conclusion \& Future work}
We explored improving the performance of neural extractive summarizers when applied to discussion threads by attending to the beginning of the text (i.e. initial comment/post) through a bidirectional attention mechanism. We showed that attending to the beginning of the text, improved ROUGE scores of different models and different variants of these models. We also showed the applicability of using a recent sentence classification model (SiATL) for extractive summarization and introduced a new SOTA ROUGE score on the trip advisor forum discussion dataset. Additionally, we showed that attending to the beginning of the text is not limited to datasets in the form of discussion threads. We showed that it is transferable to more generic forms of text, in which we can attend to the first N sentences of the text, similar to attending to the initial post/comment in discussion threads. Lastly, we showed that the utility of attending to the beginning is constant, regardless of the used model or dataset. Integrating bidirectional attention always introduces an improvement in ROUGE scores.
Future plans include trying more generic datasets such as news, to further verify the utility of attending to the beginning. Further experimenting with the SiATL model on other datasets, as it showed promising results when used as extractive summarizer. We also plan to try extending the SiATL model with a post-processing step to enforce more control over the output length. We also plan to try different values for $N$, the number of sentences as initial part from generic documents.

\bibliographystyle{aaai}
\bibliography{aaai}

\begin{thebibliography}{}

\bibitem[\protect\citeauthoryear{Bahdanau, Cho, and
  Bengio}{2014}]{bahdanau2014neural}
Bahdanau, D.; Cho, K.; and Bengio, Y.
\newblock 2014.
\newblock Neural machine translation by jointly learning to align and
  translate.
\newblock {\em arXiv preprint arXiv:1409.0473}.

\bibitem[\protect\citeauthoryear{Chronopoulou, Baziotis, and
  Potamianos}{2019}]{chronopoulou2019embarrassingly}
Chronopoulou, A.; Baziotis, C.; and Potamianos, A.
\newblock 2019.
\newblock An embarrassingly simple approach for transfer learning from
  pretrained language models.
\newblock In {\em Proc. of the 2019 Conference of NAACL-HLT, Volume 1 (Long and
  Short Papers)},  2089--2095.

\bibitem[\protect\citeauthoryear{Devlin \bgroup et al\mbox.\egroup
  }{2018}]{devlin2018bert}
Devlin, J.; Chang, M.-W.; Lee, K.; and Toutanova, K.
\newblock 2018.
\newblock Bert: Pre-training of deep bidirectional transformers for language
  understanding.
\newblock {\em arXiv preprint arXiv:1810.04805}.

\bibitem[\protect\citeauthoryear{Gehrmann, Deng, and
  Rush}{2018}]{gehrmann2018bottom}
Gehrmann, S.; Deng, Y.; and Rush, A.
\newblock 2018.
\newblock Bottom-up abstractive summarization.
\newblock In {\em Proc. of the 2018 Conference on EMNLP},  4098--4109.

\bibitem[\protect\citeauthoryear{Jha \bgroup et al\mbox.\egroup
  }{2020}]{jha2020artemis}
Jha, R.; Bi, K.; Li, Y.; Pakdaman, M.; Celikyilmaz, A.; Zhibodev, I.; and
  McDonald, K.
\newblock 2020.
\newblock Artemis: A novel annotation methodology for indicative single
  document summarization.
\newblock {\em arXiv preprint arXiv:2005.02146}.

\bibitem[\protect\citeauthoryear{Jung \bgroup et al\mbox.\egroup
  }{2019}]{jung2019earlier}
Jung, T.; Kang, D.; Mentch, L.; and Hovy, E.
\newblock 2019.
\newblock Earlier isn’t always better: Sub-aspect analysis on corpus and
  system biases in summarization.
\newblock In {\em Proc. of the 2019 Conference on EMNLP and the 9th IJCNLP},
  3315--3326.

\bibitem[\protect\citeauthoryear{Li \bgroup et al\mbox.\egroup
  }{2019}]{li-etal-2019-keep}
Li, M.; Zhang, L.; Ji, H.; and Radke, R.~J.
\newblock 2019.
\newblock Keep meeting summaries on topic: Abstractive multi-modal meeting
  summarization.
\newblock In {\em Proc. of the 57th Annual Meeting of ACL},  2190--2196.
\newblock Florence, Italy: ACL.

\bibitem[\protect\citeauthoryear{Li, Li, and Zong}{2019}]{li2019towards}
Li, J.; Li, H.; and Zong, C.
\newblock 2019.
\newblock Towards personalized review summarization via user-aware sequence
  network.
\newblock In {\em Proc. of the AAAI Conference on Artificial Intelligence},
  volume~33,  6690--6697.

\bibitem[\protect\citeauthoryear{Lin}{2004}]{lin2004rouge}
Lin, C.-Y.
\newblock 2004.
\newblock Rouge: A package for automatic evaluation of summaries.
\newblock {\em Text Summarization Branches Out}.

\bibitem[\protect\citeauthoryear{Liu and Lapata}{2019}]{liu2019text}
Liu, Y., and Lapata, M.
\newblock 2019.
\newblock Text summarization with pretrained encoders.
\newblock In {\em Proc. of the 2019 Conference on Empirical Methods in Natural
  Language Processing and the 9th International Joint Conference on Natural
  Language Processing (EMNLP-IJCNLP)},  3721--3731.

\bibitem[\protect\citeauthoryear{Luo, Liu, and Litman}{2016}]{luo2016improved}
Luo, W.; Liu, F.; and Litman, D.
\newblock 2016.
\newblock An improved phrase-based approach to annotating and summarizing
  student course responses.
\newblock In {\em Proc. of COLING 2016, the 26th International Conference on
  Computational Linguistics: Technical Papers},  53--63.

\bibitem[\protect\citeauthoryear{Nallapati, Zhai, and
  Zhou}{2017}]{nallapati2017summarunner}
Nallapati, R.; Zhai, F.; and Zhou, B.
\newblock 2017.
\newblock Summarunner: A recurrent neural network based sequence model for
  extractive summarization of documents.
\newblock In {\em Thirty-First AAAI Conference on Artificial Intelligence}.

\bibitem[\protect\citeauthoryear{Paulus, Xiong, and
  Socher}{2018}]{paulus2017deep}
Paulus, R.; Xiong, C.; and Socher, R.
\newblock 2018.
\newblock A deep reinforced model for abstractive summarization.
\newblock In {\em International Conference on Learning Representations}.

\bibitem[\protect\citeauthoryear{Rose \bgroup et al\mbox.\egroup
  }{2010}]{rose2010automatic}
Rose, S.; Engel, D.; Cramer, N.; and Cowley, W.
\newblock 2010.
\newblock Automatic keyword extraction from individual documents.
\newblock {\em Text mining: applications and theory}  1--20.

\bibitem[\protect\citeauthoryear{See, Liu, and Manning}{2017}]{see2017get}
See, A.; Liu, P.~J.; and Manning, C.~D.
\newblock 2017.
\newblock Get to the point: Summarization with pointer-generator networks.
\newblock In {\em Proc. of the 55th Annual Meeting of the ACL (Volume 1: Long
  Papers)}, volume~1,  1073--1083.

\bibitem[\protect\citeauthoryear{Seo \bgroup et al\mbox.\egroup
  }{2017}]{seo2016bidirectional}
Seo, M.; Kembhavi, A.; Farhadi, A.; and Hajishirzi, H.
\newblock 2017.
\newblock Bidirectional attention flow for machine comprehension.
\newblock In {\em Proc. of the 5th International Conference on Learning
  resentations (ICLR)}.

\bibitem[\protect\citeauthoryear{Sutskever, Vinyals, and
  Le}{2014}]{sutskever2014sequence}
Sutskever, I.; Vinyals, O.; and Le, Q.~V.
\newblock 2014.
\newblock Sequence to sequence learning with neural networks.
\newblock In {\em Advances in neural information processing systems},
  3104--3112.

\bibitem[\protect\citeauthoryear{Tarnpradab, Liu, and
  Hua}{2017}]{tarnpradab2017toward}
Tarnpradab, S.; Liu, F.; and Hua, K.~A.
\newblock 2017.
\newblock Toward extractive summarization of online forum discussions via
  hierarchical attention networks.
\newblock In {\em The Thirtieth International Flairs Conference}.

\bibitem[\protect\citeauthoryear{Wang, Quan, and Wang}{2019}]{wang2019biset}
Wang, K.; Quan, X.; and Wang, R.
\newblock 2019.
\newblock Biset: Bi-directional selective encoding with template for
  abstractive summarization.
\newblock {\em arXiv preprint arXiv:1906.05012}.

\bibitem[\protect\citeauthoryear{Xiao and Carenini}{2019}]{xiao2019extractive}
Xiao, W., and Carenini, G.
\newblock 2019.
\newblock Extractive summarization of long documents by combining global and
  local context.
\newblock {\em arXiv preprint arXiv:1909.08089}.

\bibitem[\protect\citeauthoryear{Yang \bgroup et al\mbox.\egroup
  }{2018}]{yang2018aspect}
Yang, M.; Qu, Q.; Shen, Y.; Liu, Q.; Zhao, W.; and Zhu, J.
\newblock 2018.
\newblock Aspect and sentiment aware abstractive review summarization.
\newblock In {\em Proc. of the 27th international conference on computational
  linguistics},  1110--1120.

\end{thebibliography}

\end{document}